\documentclass[runningheads]{llncs}
\usepackage[T1]{fontenc}
\usepackage{hyperref}
\usepackage{url}
\usepackage{graphicx}
\usepackage{booktabs}
\usepackage{subcaption}
\usepackage[numbers]{natbib}
\usepackage{amsmath}
\usepackage{graphicx}
\usepackage{color}

\begin{document}
\title{Learning Diachronic Representations of Ancient Greek Letterforms}
\titlerunning{Diachronic Greek Letterforms}
\author{
\textbf{John Pavlopoulos}$^{1,2}$\thanks{JP undertook the experiments and wrote the first draft; MK, GDG, IMS oversaw experiments and co-authored. LF co-authored (esp. \S4 and \S6) and provided Hell-Char and expertise. PP and HE co-authored and provided expertise. AP provided MedChar. SB and DV assisted with experiments and conceptualisation.}, 
\textbf{Spyros Barbakos}$^{2,3}$, 
\textbf{Lavinia Ferretti}$^{4,5}$,
\textbf{Dionysis Voulgarakis}$^{2}$, 
\textbf{Asimina Paparrigopoulou}$^{6}$, 
\textbf{Maria Konstantinidou}$^{6}$
\textbf{Giuseppe De Gregorio}$^{5,7}$, 
\textbf{Isabelle Marthot-Santaniello}$^{5}$, 
\textbf{Paraskevi Platanou}$^{1,2}$,
\textbf{Holger Essler}$^{8}$}

\institute{
Athens University of Economics and Business, Greece \email{ipavlopoulos@aueb.gr} \and
Archimedes, Athena Research Center, Greece\and
Department of Computer and Systems Sciences, Stockholm University, Sweden\and
Università degli Studi di Torino, Italy\and
University of Basel, Switzerland\and
Democritus University of Thrace, Greece\and
Computer Vision Center (CVC) - Barcelona, Spain\and
Julius-Maximilians-Universität Würzburg, Germany
\\
}
\authorrunning{Pavlopoulos et al.}

\maketitle              
\begin{abstract}
Learning representations that remain robust across centuries of variation in handwriting is a key challenge in diachronic representation learning.
Taking one of the longest continuously used writing systems, ancient Greek, as a case study, we introduce three datasets for diachronic representation learning: \textbf{Hell-Char}, a curated training set spanning the 3rd–1st centuries BCE, and two evaluation sets, \textbf{PaLit-Char} (2nd–5th c. CE) and \textbf{Med-Char} (9th-14th c. CE). To address the challenges of symbolic variation, scarce data, and systematic degradation, we propose: a \emph{similarity-weighted supervised contrastive loss} that biases embeddings using dynamically estimated inter-class similarities, and a \emph{lacuna-driven augmentation} scheme that simulates realistic manuscript corruptions. Trained with these strategies, both a lightweight CNN and a pretrained ResNet achieve strong recognition performance and produce embeddings that more coherently separate character classes than PCA or generic pretrained models. These embeddings enable clustering, identification of stylistic subgroups, and construction of prototype images that visualize diachronic evolution and transitional letterforms. Our results demonstrate that respecting intrinsic inter-letter relationships and augmenting with domain-informed corruptions yield robust, interpretable representations, offering a transferable paradigm for representation learning under scarce, temporally evolving, and noisy conditions. Code and data available at: \href{https://github.com/ipavlopoulos/diachronic-greek-letterforms}{https://github.com/ipavlopoulos/diachronic-greek-letterforms}.

\end{abstract}

\setcounter{footnote}{0}
\section{Introduction}
Paleographic analysis of historical scripts requires robust automated character representation, a challenge that remains particularly acute for scripts such as ancient Greek. Greek handwriting spans over two and a half millennia, encompassing formal literary hands and highly cursive scripts, with substantial variation in stroke shape, scale, slant, and contextual noise~\citep{cavallo_greek_2009, crisci_scrittura_2011, irigoin_alphabet_1990, bianconi_greek_2015}. Material degradation and heterogeneous digitization practices further compound these challenges, introducing ambiguities that complicate segmentation, feature extraction, and character recognition and classification, especially under limited and imbalanced datasets. Although sometimes considered a low-level task, automated character representation has a significant impact for broader paleographic analysis, supporting text-image alignment, semi-automatic transcription, and tasks such as script typology, dating, and scribal attribution. 

Existing document analysis and recognition methods typically assume stable character morphology and sufficient training data. These assumptions break down in historical scripts, where letterforms evolve gradually across centuries and exhibit systematic structural drift. Hence, standard transfer and contrastive learning approaches do not explicitly model structured inter-class similarity, treating visually related but distinct letterforms as equally dissimilar negatives.

This study addresses this challenge by learning robust character representations, with diachronic generalization serving as the central test case. We focus on the evolution of ancient Greek letters through a representation learning framework that incorporates paleographic knowledge. 
We propose two domain-driven methodological innovations: a \emph{lacuna-driven augmentation} (LF) that simulates realistic manuscript degradations, and a \emph{similarity-weighted} supervised contrastive loss (DSCL) that reweights negative pairs according to dynamically estimated inter-class similarities. 
LF targets the non-rectangular loss patterns produced by manuscript damage, while DSCL targets morphologically confusable letter classes that should not be treated as uniformly distant negatives.
We evaluate their impact on both recognition performance and representation structure. Confusion matrices reveal systematically easy and difficult letters, while clustering analyses on the learned embeddings uncover stylistic subgroups and visually coherent forms. Prototype visualizations per letter–century further enable quantitative and interpretable analysis of diachronic evolution. Compared to raw pixels, PCA, or generic pre-trained features, our embeddings yield more coherent and discriminative representations of historical Greek handwriting.

We summarize our contributions as the following four key points:

\noindent\textbf{1. We propose a representation learning objective} that, unlike standard SCL or hard-negative mining, explicitly models inter-class similarity structure, preventing repulsion between inherently similar graphemes.

\noindent\textbf{2. We introduce a domain-informed augmentation scheme} that simulates manuscript degradations (lacunae) more faithfully, increasing robustness to missing or corrupted strokes.

\noindent\textbf{3. We introduce historical Greek handwriting datasets} spanning from the 3rd century BCE to the 14th century CE: Hell-Char (3rd–1st BCE), derived from Hell-Date for character-level training and benchmarking; and two newly compiled evaluation datasets, PaLit-Char (2nd–5th CE) and Med-Char (9th–14th CE) for testing generalization across temporal shifts.

\noindent\textbf{4. We conduct computational paleographic analyses} using CNN-derived embeddings. We perform clustering, silhouette-based subgroup detection, and prototype visualization per letter--century, providing interpretable insights into diachronic variation and scribal conventions.\footnote{Code and data are available at: https://github.com/ipavlopoulos/diachronic-greek-letterforms.}

\section{Related Work}\label{sec:literature}
We are not aware of any other study in the literature that 
analyses the diachronic evolution of Greek handwritten letters from Antiquity to pre-modern times using machine learning. However, we acknowledge the existence of related fields, such as optical character recognition (OCR), and of other investigations on Greek papyri at the character level, which we discuss next.   

\paragraph{OCR.} Early OCR approaches relied on manual feature extraction methods, such as zoning, projection histograms, and contour profiling, to distinguish between characters. A comprehensive survey emphasized the importance of these handcrafted features in OCR \cite{Trier_1996}, while in \cite{He_2016} a grapheme-based feature extraction system was introduced that modelled diachronic variations while incorporating textual features.
The advent of deep learning has further transformed the field. In \cite{LeCun_1998}, the authors demonstrated the effectiveness of convolutional neural networks (CNNs) in classifying handwritten digits, laying the groundwork for modern neural approaches in character recognition. Autoencoders \cite{Hinton_2006} and contrastive learning \cite{Chen_2020} have gained traction in unsupervised learning, enabling models to learn meaningful representations of handwriting directly from data, without the need for manual feature engineering.
Building on these advances, several of the latter works have examined deep learning for feature analysis of some aspects of ancient Greek handwritings. In \cite{Marthot_2023}, the authors addressed the issue of clustering historical handwriting by similarity with no metadata explicitly indicating date or style. Their method strongly focuses on character, employing a SimSiam neural network to quantify similarity between images of single Greek letters (Alpha, Epsilon, and Mu) from different manuscripts. Their observations of stylistic similarity are useful to paleographers as they situate manuscripts within an integrated network and disclose subtle micro-phenomena of similarity.

\paragraph{CNNs.} CNNs have been applied to OCR-extracted text \cite{Li_2015}, combining visual and textual features to improve dating accuracy. However, their approach assumes the availability of high-quality (historical yet printed) data conducive to accurate OCR results, an assumption that often fails in the context of historical documents such as the Greek papyri addressed in our study. To tackle such challenges, an ImageNet-pretrained CNN was fine-tuned on a corpus of medieval documents, demonstrating improved performance on degraded or irregular scripts \cite{Wahlberg_2016}.
More chronology-specific, in \cite{west_2024}, a deep learning pipeline was designed for the automated dating of images of ancient Greek papyrus fragments. A multi-stage pipeline integrated handwritten text recognition (HTR) for character detection and classification, followed by distinct character-level and fragment-level dating prediction models.
Their aggregated sum of fragment-level models reaches up to 79\% accuracy in predicting two-century-broad date ranges on fragments with large numbers of characters. 
More recently, a transformer-based pipeline was proposed \cite{Boudraa_2024} that integrates classical preprocessing techniques with a fine-tuned Vision Transformer and majority voting for document dating. This study pioneers the integration of Vision Transformers in the context of historical manuscript dating, a domain where CNNs were dominating. 

\paragraph{SimCLR.} A simple yet powerful contrastive framework for representation learning was introduced in \cite{Chen_2020}. Each image is augmented twice, and the network is trained to maximize agreement between positive pairs while treating all other samples in the batch as negatives. While effective as a simple self-supervised technique at scale, SimCLR assumes that all non-matching samples are equally dissimilar. In fine-grained recognition tasks such as character classification, this uniform treatment forces visually similar but distinct classes apart (e.g., A vs. $\Lambda$), discarding useful structural information.  

\paragraph{SCL.}
In \cite{khosla2020supcon}, the authors extended SimCLR to the labelled setting by grouping all samples of the same class as positives. This produces tighter class-specific clusters. Importantly, they also showed that combining supervised contrastive embeddings with a linear classifier trained under cross-entropy further improves classification accuracy compared to cross-entropy alone. However, SCL still treats all negatives uniformly, regardless of their visual similarity to the anchor. As a result, classes with inherent affinities (e.g., letters with similar shapes) are repelled too strongly, yielding embeddings that fail to reflect natural inter-class relationships. In addition to instance discrimination, weakly SCL~\citep{zheng2021weakly}  introduced a supervised contrastive component based on weak labels derived from K-nearest neighbor graphs. Instead of treating all other samples as negatives, this approach dynamically identifies semantically similar neighbors and reweights them as positives, alleviating the class collision problem. SCL treats negatives uniformly and makes classes with inherent affinities (e.g., letters with similar shapes) to be strongly repelled. This leads to embeddings that fail to reflect natural inter-class relationships. Our study addresses this gap. 

\section{Methodology}
We analyse handwritten Greek letters across centuries using CNN-based embeddings enhanced with advanced fragmentation and SCL strategies.

\subsection{CNN Backbone and Lacunae Fragmentation}\label{ssec:lefcnn} A 2D CNN (fCNN) was suggested in \cite{pavlopoulos2024explainable} for dating images of papyrus lines, which comprised a fragmentation-based augmentation strategy. We follow a similar fragmentation-based strategy, yet our CNN differs in two ways. First, it is adjusted to operate on letters instead of text lines. Second, the fragmentation augmentation is improved so that synthetic lacunae follow their natural (curvy) shape, i.e., circular or elliptic, not square (\S\ref{ssec:settings}). The trained model produces high-dimensional embeddings $\mathbf{e} \in \mathbf{R}^D$ representing the visual structure of each letter. The base CNN architecture consists of convolutional layers to extract local stroke and shape patterns; ReLU activations for non-linearity; pooling layers to reduce spatial dimensions while preserving salient features; fully connected layers to map feature maps into the final embedding vector. These embeddings abstract style variations while preserving essential letterform characteristics. We also experiment with ResNet18 pre-trained CNN \citep{he2016deep}, the ConvNext-V2 self-supervised and globally-normalized CNN \cite{Woo_2023_CVPR}, and the ViT-S16 Vision Transformer \cite{Caron2021EmergingPI}.

\subsection{Augmentation}
\begin{figure}[ht]
    \centering
    \includegraphics[width=\textwidth]{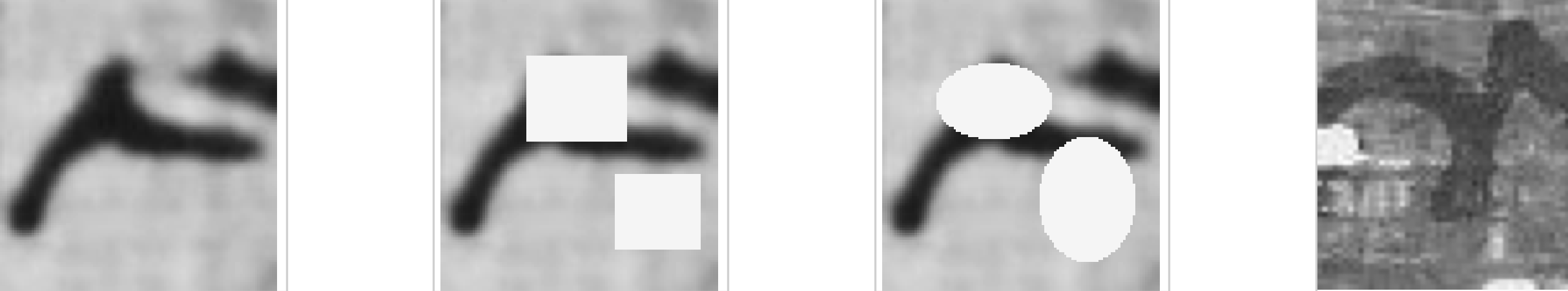}
    \caption{Alpha from Hell-Char (leftmost) with rectangular (RE, 2nd) and lacuna-inspired (LF, 3rd) masking. Alpha with naturally fragmented surface (4th).}
    \label{fig:lf-aug}
\end{figure}

Each character image is converted to grayscale, normalized, and resized to $64\times64$ pixels. To account for variability in handwriting and material degradation, we applied rotation (up to 10°), translation, resizing, color jittering, and lacunae-inspired masking (LF), which simulates missing ink or manuscript damage, improving the model’s robustness to partial character visibility (Fig.~\ref{fig:lf-aug}, rightmost).

\subsection{Similarity-Weighted Supervised Contrastive Loss} 
In addition to the standard cross-entropy loss (i.e., the supervised letter-classification objective applied to the backbone’s classification head), we train the backbone models using a SCL loss, which encourages embeddings of the same letter to cluster together while pushing apart visually dissimilar letters. Visual similarities between letters, dynamically learned,\footnote{Visual similarities could also be defined manually, but our experiments (using a prior similarity matrix based on modern letter shapes) did not lead to improvements.} are used to weight negative pairs, enabling the model to respect intrinsic inter-letter relationships. 
This contrastive loss is not computed on the classification logits, but it is applied to the intermediate feature embeddings produced by the backbone before the classification layer. Thus, the model jointly optimizes cross-entropy on the classification head and contrastive loss on the shared backbone representations.
For each anchor embedding $\mathbf{e}_i$, the loss is defined as:

\[
\mathcal{L}_i = - \frac{1}{|P(i)|} \sum_{p \in P(i)} \log \frac{\exp(\mathbf{e}_i \cdot \mathbf{e}_p / \tau)}{\sum_{a \neq i} w_{ia} \exp(\mathbf{e}_i \cdot \mathbf{e}_a / \tau)}
\]

\noindent where $P(i)$ is the set of positive samples (same class as $i$) and $\tau$ is the softmax temperature; $w_{ia} = 1 + \lambda \frac{S_{y_i, y_a}}{\bar{S}}$ is the weight for negative pair $(i,a)$; $S_{y_i, y_a}$ is the similarity between classes $y_i$ and $y_a$, dynamically computed from embeddings; $\bar{S}$ is the mean off-diagonal similarity; and $\lambda$ controls the influence of similarity weighting. This loss ensures that embeddings of the same letter cluster tightly, while visually similar letters exert weaker repulsion.

\subsection{Prototype Selection (Medoid)}\label{ssec:medoids}
For each group $(\text{letter}, \text{century})$, we select a representative \emph{medoid} embedding to serve as a prototype ($T$), defined as: $T = \arg\min_i \sum_{j=1}^{N} \big( 1 - \cos(\mathbf{e}_c, \mathbf{e}_j) \big)$, where $N$ is the number of embeddings in the group and $e_c$ is the centroid. The medoid provides a highly representative image that is robust to outliers.

\section{Dataset Development}

\subsection{Source}
The Hell-Date dataset~\citep{ferretti_hell-date_2025} comprises 194 images from 157 papyri, all written in Greek and dated between the years 310 BCE and 3 BCE. The material is particularly relevant for digital paleography and papyrological analysis due to its historical span, script diversity, and accompanying metadata. Each document in the dataset is associated with rich contextual metadata, including the date of composition, the geographical provenance, and the textual type. Of the 194 available images, 171 are annotated at the character level, forming the primary subset of character images used in this work. We used this character-level subset but filtered and restructured it for our purposes; we refer to the restructured subset as Hell-Char (see below). This is the first study to utilize the character annotations included in Hell-Date, which are further presented below.

\paragraph{The character annotations in Hell-Date.}
Twenty-nine character classes are present in the annotations of Hell-Date. In addition to the 24 standard letters of the Greek alphabet, the dataset comprises 3 archaic numeral letters (\textit{Stigma}, \textit{Qoppa}, and \textit{Sampi}). It also uses a general `symbol' category for all characters that are not alphabetic letters. Last, an `unknown' class was added for uncertain or ambiguous signs, but it remained empty.
Each character instance is also assigned a base-type (BT) tag, ranging from BT1 to BT5, which indicates its degree of preservation. BT1 is the best-preserved, but all can be useful in the future to analyze the correlation between physical degradation and recognition performance.

\subsection{The Hell-Char subset} 
\begin{figure}
    \centering
    \includegraphics[width=\linewidth]{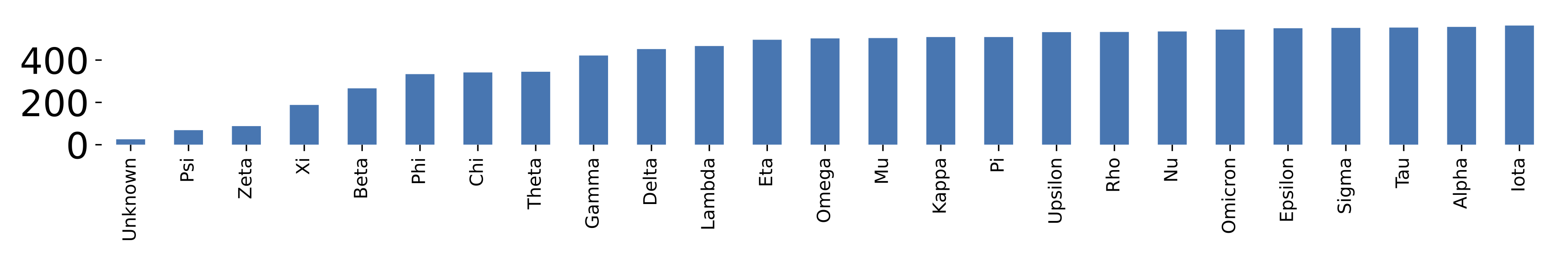}
    \caption{Letter frequency in our Hell-Char subset.}
    \label{fig:letter-dist}
\end{figure}

As said above, Hell-Char is a subset of Hell-Date annotations. The classes for archaic numerals, symbols, and unknown letters were merged into a single general non-alphabetic category labelled "Unknown", which was ignored in later stages of our work. 
We also limited our analysis to letters tagged BT1. To reduce the imbalance in character frequency and to ensure a more uniform distribution of samples across classes and documents, at most five instances per character class per papyrus were randomly selected. The resulting subset comprises 13,046 character images from 157 distinct papyri. Figure~\ref{fig:letter-dist} shows the letter frequency in Hell-Char.

\paragraph{Intrinsic challenges.} The dataset presents several intrinsic challenges. The character class distribution is still unbalanced (Figure~\ref{fig:letter-dist}) due to the large variation in letter frequency in Greek (the highest frequency is over 10 percent for Alpha, while below 0.5 percent for the three rarest letters Psi, Zeta and Xi). Besides, some glyphs are ambiguous even to experts when considered without any linguistic or chronological context. These characteristics make Hell-Char a valuable, non-trivial benchmark for evaluating character recognition in ancient scripts.

\section{Empirical Analysis}

\subsection{Experimental Settings}\label{ssec:settings}
\paragraph{The Similarity Matrix.} We re-estimate the class-similarity matrix periodically (every 3 train epochs). At each update, we pass the entire training set through the current model, compute class prototypes from the normalized embeddings, and derive the cosine similarity between prototypes. Cosine similarities are clamped to $[0,1]$ and diagonal entries are set to zero. This yields a dynamic measure of inter-class similarity that evolves with the representation space; an exponential moving average can be applied to stabilize updates. The updated matrix is then used by our DSCL, which down-weights negatives from highly similar classes and up-weights negatives from dissimilar ones.

\paragraph{Lacuna-driven Synthetic Fragmentation.}
We attempt to simulate manuscript deterioration more realistically than standard erasure augmentations by inserting irregular regions that approximate actual lacunae observed in historical documents. For each image, we sample 1–4 lacunae and each covers 2–15\% of the area to match the typical size distribution of physical papyrus damage. Lacuna geometry is obtained by drawing anisotropic ellipses whose contours are further distorted via random morphological operations (erosion or dilation), producing organic, non-rectangular shapes characteristic of flaking, humidity damage, surface wear, or other deterioration (e.g., worm holes are frequent in papyri). These lacunae are placed at random positions and the masked pixels are replaced with background values, reflecting the absence of ink/papyrus rather than additive noise. This augmentation increases robustness to fragmentary handwriting and introduces realistic variability at negligible computational cost.

\paragraph{Data Split.}
We keep 20\% of the data for testing, following a letter-based stratified split. Although we acknowledge that this strategy allows a scribe-based leakage, the selected approach better fits the scope of this work (see \S\ref{sec:discussion}). 

\subsection{Letter Recognition}
\label{letter_recognition}
Table~\ref{tab:clf-results} shows the performance of backbones when we add our LF and DSCL. 
\begin{table}[h]
\centering\caption{Classification performance on Hell-Char (sorted) of fCNN \citep{pavlopoulos2024explainable} and ResNet18 \citep{He_2016}, pre-trained (PT) and/or fine-tuned (FT), when we add: our SCL with dynamically-learned weights (DSCL), and fragmentation-based augmentation: i.e., none (-), random erasure (RE), or our lacuna-driven (LF).}
\begin{tabular}{l c c c c}
\toprule
\textbf{Model} & Fragmentation & Contrastive Loss & \textbf{Accuracy} & \textbf{F1} \\
\midrule
fCNN & - & - & 0.742 & 0.74\\ \midrule
fCNN & RE & - & 0.768 & 0.75\\ 
fCNN & LF & - & 0.782 & 0.77\\ \midrule
ResNet18-FT & - & - & 0.788 & 0.74\\ 
ResNet18-PT+FT & - & - & 0.801 &  0.79\\ \midrule
ResNet18-PT+FT & - & SCL & 0.818 & 0.81\\\midrule
fCNN (ours) & LF & DSCL & 0.803 & 0.80\\ 
ResNet18-PT+FT (ours) & LF & DSCL & 0.831 & \textbf{0.83}\\
\bottomrule
\end{tabular}
\label{tab:clf-results}
\end{table}
The vanilla fCNN \cite{pavlopoulos2024explainable} without any fragmentation, achieves an Accuracy of 74\%. F1 is exactly the same, indicating the balanced performance across letters despite the class imbalance (Figure~\ref{fig:letter-dist}). Random erasure (RE) \cite{pavlopoulos2024explainable}, shown in Fig.~\ref{fig:lf-aug}, leads to improvements in both metrics. Our LF, however, outperforms both. The architecture of ResNet18 \citep{He_2016} performs worse in F1 (on par in Accuracy) when trained from scratch, and improves when pre-trained. Standard SCL improves further ResNet18-PT+FT, which is outperformed only by our LF+DSCL version (0.83). When we enhance fCNN with our LF and DSCL, it outperforms the pre-trained ResNet18. 
Preliminary experiments show that LF and DSCL also improve the classification performance of ViT-16S and ConvNeXt-V2, but an analytical investigation of the gains across backbones is left for future work.

\subsection{Letter Image Clustering}
\label{letter_image_clustering}
Our previous experiments showed that CNN-based embeddings can be used to represent letters. 
To assess the quality of the resulting embeddings, we compared them against baseline features, then feeding algorithms that should cluster images of the same letter into subcategories. 
We also engineered features, based on Otsu's method~\citep{otsu_1979}, a widely used adaptive thresholding technique, and principal component analysis (PCA) \citep{Pearson_1901}, keeping as many dimensions as add up to 90\% of the original information (i.e., 500).    
Characters have consistent alignment and size, hence pixel-based variance captured by PCA can correspond to meaningful features of the characters (e.g., strokes and overall shape). Although it destroys 2D structure (edges, texture) and does not focus on separability, PCA applied to the raw input is a simple preprocessing baseline that is complementary to CNN features that preserve local structure.
\begin{table}[ht]
\centering\caption{Clustering performance on Hell-Char using different embeddings and clustering algorithms, sorted by performance across algorithms.}
\begin{tabular}{p{6.2cm}ccccccc}
\toprule
 & \multicolumn{2}{c}{k-means} & \multicolumn{2}{c}{Spectral} & \multicolumn{2}{c}{AH} \\
\cmidrule(lr){2-3} \cmidrule(lr){4-5} \cmidrule(lr){6-7}
Embedding & NMI & ARI & NMI & ARI & NMI & ARI \\
\midrule
\textsc{ResNet18+LF+DSCL} & \textbf{0.768} & \textbf{0.690} & \textbf{0.765} & \textbf{0.683} & \textbf{0.760} & \textbf{0.674} \\
fCNN+LF+DSCL & 0.428 & 0.189 & 0.631 & 0.442 & 0.544 & 0.292 \\
\textsc{ResNet18+PT+FT} & 0.480 & 0.257 & 0.487 & 0.225 & 0.464 & 0.197 \\
Otsu+PCA & 0.318 & 0.152 & 0.382 & 0.176 & 0.356 & 0.168 \\
\textsc{ResNet18+PT} & 0.067 & 0.010 & 0.094 & 0.015 & 0.073 & 0.008 \\
\bottomrule
\end{tabular}
\label{tab:clustering}
\end{table}
The empirical results, shown in Table~\ref{tab:clustering}, underscore the importance of task-specific embeddings in historical handwriting clustering. ResNet18, fine-tuned and enhanced with LF and DSCL, consistently outperforms both Otsu+PCA and a pre-trained ResNet18, achieving markedly higher agreement with paleographic labels across all metrics. 
Otsu+PCA, though superior to raw pretrained features, lags far behind, while ResNet18 fails entirely, with near-random partitions. The stark contrast highlights two key findings: (i) general-purpose CNN features trained on \href{https://www.image-net.org}{ImageNet}-style photographs are far from isolated historical character images and do not transfer directly to paleographic tasks, and (ii) the manifold structure of handwritten letter embeddings is not captured adequately by centroid-based partitioning. Together, these results validate the need for domain-tailored architectures and manifold-aware clustering to recover meaningful structure in diachronic handwriting data.

\subsection{Pattern Recognition: Revealing Letter Forms}
\begin{figure}[ht]
    \centering
    \includegraphics[width=.5\textwidth]{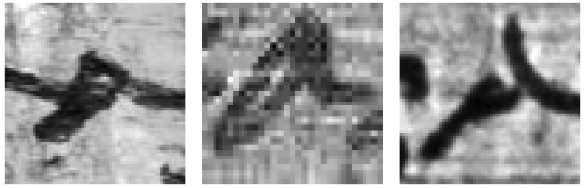}
    \caption{Representative forms of letter Alpha using cluster medoids (\S\ref{ssec:medoids}).
    }
    \label{fig:letter_forms}
\end{figure}

Using Spectral Clustering on the embeddings of our best performing CNN (i.e., \textsc{ResNet18+lf+dscl}; see Table~\ref{tab:clf-results}), we applied the Silhouette method \citep{schubert2023stop} to detect the optimal number of clusters per letter. For each letter, we varied the number of clusters and retained the configuration with the highest Silhouette score \citep{rousseeuw1987silhouettes}. For Alpha, the optimal number exceeded the two clusters indicating multiple distinct forms,\footnote{Silhouette scores cannot be computed for a single cluster; hence the minimum number considered was two.} shown in Figure~\ref{fig:letter_forms}.
Clustering letter forms into subtypes is a hard and unsolved task in paleography. The network's results may, in some cases, point to useful characteristics. In the case of Alpha, the three images seem to represent different characteristics: the first alpha is filled-in (no empty space in its centre), circular and ligatured to the left; the second one is circular but not ligatured; the third one is angular. This partition is coherent from a paleographical point of view. 

\section{Out of Temporal Distribution Application}\label{sec:ood}

Our backbone CNNs are trained on letter images from papyri of the last three centuries BCE. In this period, the epigraphic letter forms (close to our modern capital letters) start to be modified with increasing cursivity, driven by the practical demands of faster writing. This increase in cursivity continues over the following centuries and constantly deforms letter shapes; however, epigraphic letter forms are maintained, especially for calligraphic writing styles called capital (or uncial) bookhands. In parallel, a calligraphic stylization of cursive forms gradually develops until it reached the so-called state of ``minuscule script'' in the 9th c. While many minuscule letterforms remain visually close to their capital ancestors, others diverge significantly (e.g. Beta, Mu, Gamma, and Delta). During the following centuries, uncial and minuscule calligraphic forms continued to coexist, sometimes within the same manuscript and even within a single word.

\subsection{Evaluation Dataset Development}
\paragraph{PaLit-Char: Majuscule Literary Papyri.}
To evaluate how well the model generalizes to letterforms close in time to the training data, we constructed the \textbf{PaLit-Char} test set. It is a fully balanced dataset containing 384 images (4 specimens × 24 letters × 4 centuries) spanning the 2nd–5th CE. Images were drawn from securely dated literary papyri in the PaLit dataset \citep{pavlopoulos2024explainable}; for the 5th century, where securely dated material is scarce, 48 images were taken from an additional, paleographically dated manuscript. While Hell-Char covers cursive handwriting from the last three centuries BCE, PaLit-Char extends into the early centuries CE and covers calligraphic writing, offering both chronological continuity and stylistic diversity. This allows us to test whether features learned on late Hellenistic cursive letters transfer to Roman-period bookhands that retain strong ties to their predecessors but already display variation.
 
\paragraph{Med-Char: Medieval Minuscule Manuscripts.}
With the historical evolution described above in mind—and having first tested the recognition performance of our network on the chronologically close PaLit-Char—we proceed to test its ability to recognize letterforms from medieval minuscule manuscripts. This evaluates both the generalizability of learned features across paleographic periods and the limits of shape-based classification given the diachronic script variation. To assess this hypothesis, we compiled a dataset of 574 letter images from manuscripts dated between 835 and 1378 CE, a much later period. We used 24 images per letter, opting for balance across the centuries in that period,\footnote{We include 24 random instances of each letter per century from multiple manuscripts. The rarest letter, Psi, has only 22 occurrences.} and using the best performing ResNet18, enhanced with our LF and DSCL, to classify each image. We call this evaluation dataset \textbf{Med-Char}. Contrary to our training set and the PaLit-Char test set, which contain capital or cursive letters (upper case), Med-Char is a Byzantine minuscule letter (lower case) dataset. This choice is deliberate: minuscule script is historically derived from majuscule but exhibits substantial graphic divergence, with some letters retaining visual continuity and others undergoing radical transformation. Testing on Med-Char, hence, allows us to probe the limits of the learned representations under extreme diachronic and stylistic shift, providing a benchmark for cross-period generalization.

\subsection{Experimental Analysis}

\paragraph{Closer in Time.}
On PaLit-Char (Table~\ref{tab:diachronic_summary}), \textsc{ResNet18+lf+dscl} achieves an Accuracy and F1 of 0.84, very close to the results of Hell-Char.
Although F1 decreased for specific letters (e.g., Phi, Pi, Psi), it improved for others (e.g., Alpha and Zeta). The calligraphic nature (regular, standardized, legible) of PaLit characters can explain this increase.
\begin{table}[h]
\centering
\caption{Diachronic generalization of \textsc{ResNet18+lf+dscl} across datasets.}\label{tab:diachronic_summary}
\begin{tabular}{lccc}
\hline
\textbf{Test Set} & \textbf{Period} & \textbf{Accuracy} & \textbf{F1-score} \\
\hline
Hell-Char  & 3rd--1st c. BCE & 0.83 & 0.82 \\
PaLit-Char & 2nd--5th c. CE  & 0.84 & 0.84 \\
Med-Char   & 9th--14th c. CE & 0.45 & 0.42 \\
\hline
\end{tabular}
\end{table}

\paragraph{Far Away in Time.}
The same model reaches 0.45 Accuracy on Med-Char with highly uneven per-class performance. Chi, Lambda and Iota are recognised best (F1 0.87, 0.80, 0.77), as their capital, cursive and minuscule shapes remain similar. In contrast, Alpha, Gamma and Upsilon collapse to zero F1, reflecting systematic confusion with visually similar shapes and high diachronic variability that undermines generalization. Gamma, for instance, undergoes a strong visual evolution: in Hell-Char, its shape is still close to epigraphic Gamma $\Gamma$, whereas in Med-Char, it is close to minuscule Gamma $\gamma$, a shape that resembles epigraphic Upsilon of $\Upsilon$ in Hell-Char.
The remaining letters trade Precision against Recall: Omicron, Tau, and Kappa are over-predicted (Recall 1.00, 0.96, 0.88, but Precision 0.23, 0.48, 0.41), whereas Psi is under-predicted (Precision 0.90, Recall 0.41), and Delta is almost never predicted, but in the rare cases it is, it is correct. 
\begin{figure}
    \centering
    \includegraphics[width=\textwidth,trim={0 0 0 25},clip]{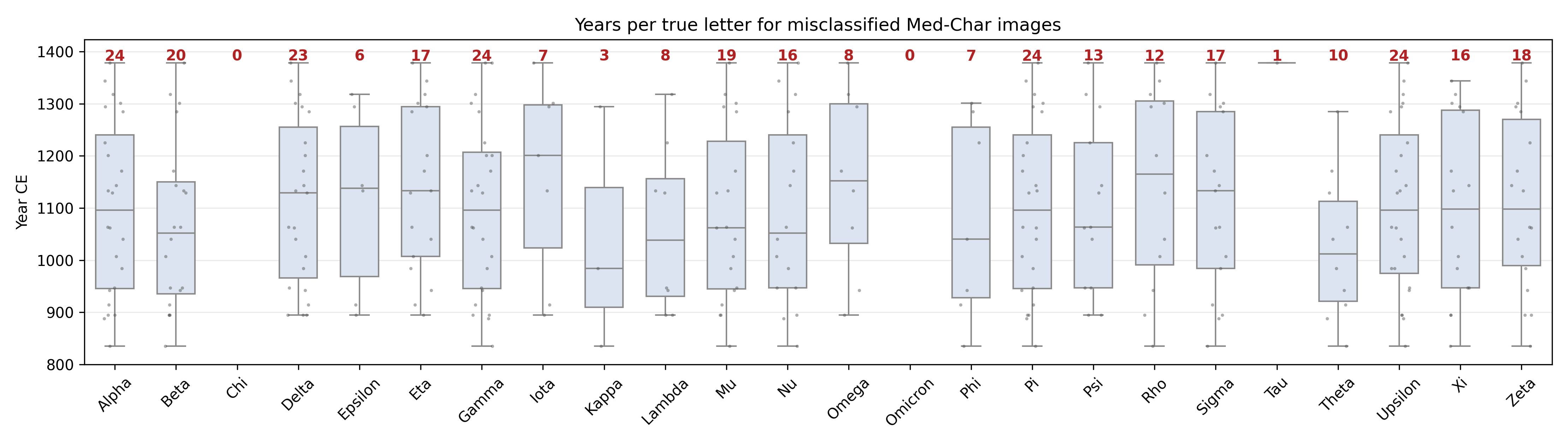}
    \caption{Boxplot of years per letter for misclassified out-of-distribution images of Med-Char. Red numerals indicate the number of mistakes.}
    \label{fig:temporal-distribution}
\end{figure}
As can be seen in Figure~\ref{fig:temporal-distribution}, misclassification patterns are temporally structured. Best-recognised Chi and over-predicted Omicron have no (Recall) errors, and Tau only a single one (near 1378 CE). For most letters the errors concentrate in the 11th--12th centuries (median around 1050--1130 CE). The earliest confusions involve Kappa (around 985 CE) and the latest Iota (around 1200 CE), implying that historical morphological shifts exert non-uniform effects on recognition difficulty. Noteworthy is the fact that fine-tuning on PaLit-Char and inferring on Med-Char brings only modest gain, from 0.45 to 0.48 Accuracy. This transfer remains difficult because Roman and late-antique majuscule bookhands differ structurally from medieval minuscule, and the intermediate cursive stages that connect them are absent from the training data.

\begin{figure}[h]
    \centering
    \includegraphics[width=\textwidth,trim={0 0 0 20},clip]{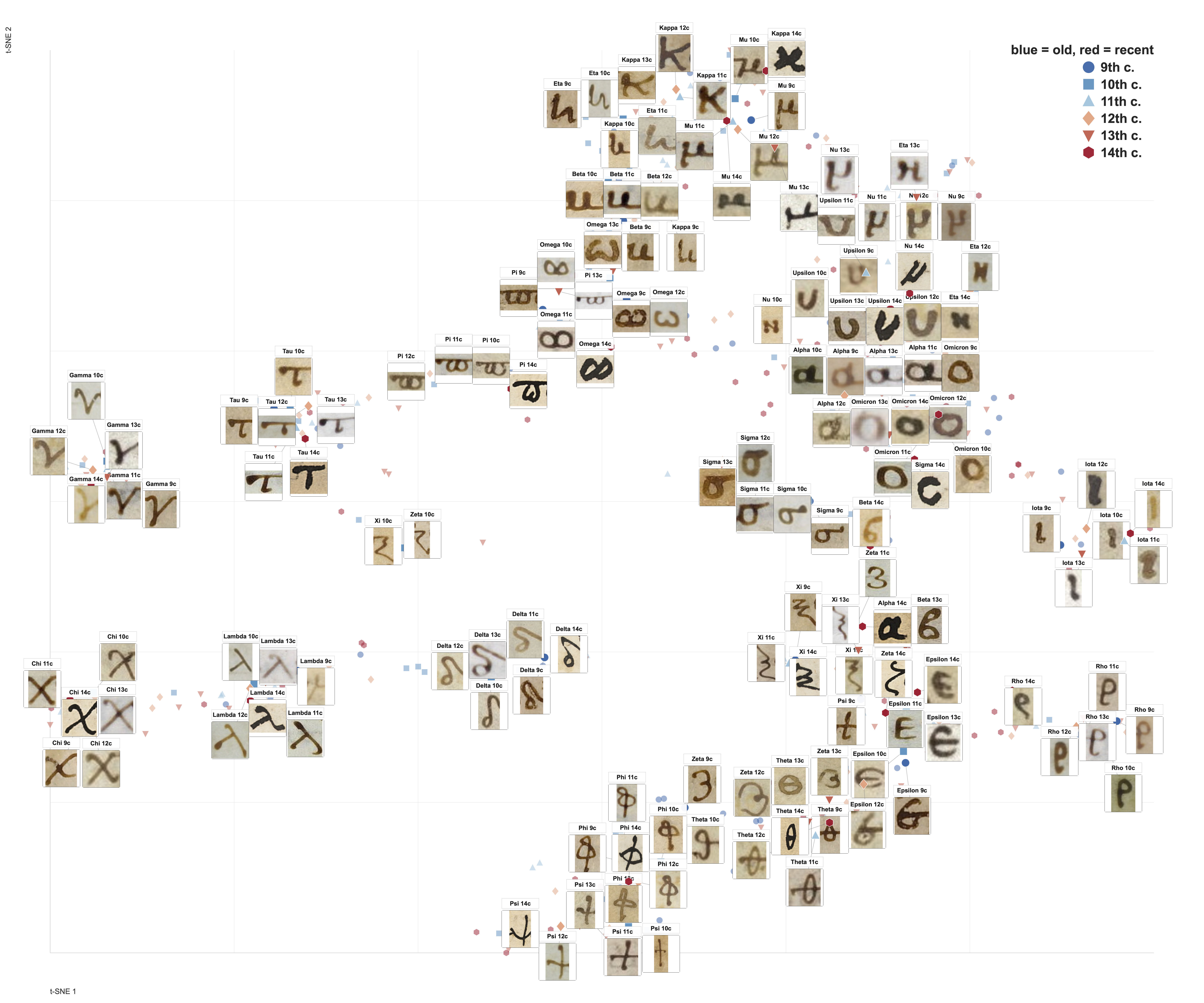}
    \caption{t-SNE plot of \textsc{ResNet18+lf+dscl} embeddings on Med-Char. Points are coloured by century (blue for older, red for more recent). One prototype image per letter-century group is shown, selected as the sample closest to the group centroid, to visualize graphemic clustering and local diachronic variation.}
    \label{fig:letter-century-plot}
\end{figure}

\paragraph{Letter-Century Clusters.} 
Figure~\ref{fig:letter-century-plot} illustrates a two-dimensional t-SNE projection of the \textsc{ResNet18+lf+dscl} embeddings, where each point corresponds to an image patch representing a handwritten Greek Med-Char character. To reduce clutter and improve interpretability, instead of showing all individual samples, one prototype image per letter–century pair is overlaid: the prototype is chosen as the sample closest to the centroid of its group in the t-SNE space, thus representing the most ``typical'' example of that cluster. The resulting map highlights how temporal and graphemic factors shape the embedding space. 
Within the clusters related to one character, overlapping or diffuse areas indicate stylistic continuities or transitional forms between centuries, whereas sharp separations reveal periods of stronger diachronic variation. This approach provides an interpretable way of assessing the alignment between automated embeddings and paleographic expectations, enabling both qualitative validation of the clustering behavior and the identification of anomalies or particularly distinctive exemplars.

\noindent\textbf{Distinctively Isolated Letters.} Several letters that obtained high F1 scores (e.g.,  Iota $\iota$, Lambda $\lambda$, Chi $\chi$) form distinct peripheral clusters, consistent with their relatively stable shapes across the training material. 
Gamma also forms an isolated cluster, but its form varies greatly from Hellenistic to Medieval times. 
This strong diachronic shift can explain why the model fails to recover Gamma in Med-Char: true medieval Gamma examples no longer match the Gamma forms learned from Hell-Char, leading to zero Precision and Recall for the class.

\noindent\textbf{Visible Clusters: The cases of Zeta and Beta.} 
Multiple forms of Zeta are visible, including examples whose shape resembles a 3 and others closer to modern-day $\zeta$, often near letters with related stroke structure such as Theta or Xi. This distinction points to one reason for confusion in recognizing Zeta: its ``3-looking'' shape is absent from the training data and therefore, cannot be identified. The same is true for Beta: 
its B-shaped and minuscule, u-shaped forms occupy different neighbourhoods of the embedding space, with the latter appearing near other minuscule u-shaped letters such as Kappa, Mu and Eta. This u-shaped Beta, absent from Hell-Char, explains its very low Recall.

\noindent\textbf{Typical Medieval Forms} A broad region of the graph groups typical Medieval letter forms based on successions of `o' and `u' shapes. These shapes are quite different from Hell-Char letters shapes, and indeed some of the letters represented here (Omega, Beta, Kappa, Mu, Nu, Upsilon) do not belong to the top-performing ones. Kappa and Nu achieved an F1 of 0.56 and 0.47 respectively, which can be explained by their mixing Medieval, minuscule, u-shaped forms with older, capital forms already attested in Hell-Char. These older K and N forms appear near the edges of this broader minuscule-form neighborhood.

\section{Discussion}\label{sec:discussion}

\paragraph{Novel Methodological Contributions.} Our core innovations lie in the LF augmentation scheme and DSCL similarity-weighted loss. The empirical results clearly demonstrate that these methods significantly outperform standard CNN baselines and generic pre-trained models, which often fail to transfer to the variability of paleographic data. Erasing input as augmentation in image classification is not new \cite{zhong2020random}, and it has been shown to be particularly useful for papyri, which are often fragmented. \textbf{Our presented LF augmentation} is closer in nature to the real fragments compared to RE (see Figure~\ref{fig:lf-aug}),
while \textbf{our proposed DSCL} 
yields embeddings that reflect natural inter-class relationships. 
By contrast, standard SCL treats negatives uniformly, making classes with inherent affinities (e.g., letters with similar shapes, such as Psi-Phi) to be strongly repelled. 

\paragraph{Scribe Leakage and Diachronic Generalization.} We assess the robustness of our representations by testing the model on chronologically distinct datasets (Table~\ref{tab:diachronic_summary}). The stable performance on PaLit-Char (0.84 F1) confirms that the model captures structural letterforms rather than individual handwriting styles, effectively mitigating the risk of scribe leakage. The performance drop observed on Med-Char (0.42 F1) further reinforces this defense: it reveals that the model is sensitive to the radical morphological shift from Hellenistic cursive to Byzantine minuscule (e.g., the evolution of $\Gamma$ into $\gamma$). Our error analysis shows that the model, lacking intermediary Roman-period cursive data, maps unknown minuscule shapes to the closest visual proxies, such as confusing Med-Char Gamma with Upsilon. This confirms that our framework successfully learns transferable paleographic features that follow historical evolutionary paths rather than memorizing specific training samples. While the low sample density (max 120 characters per papyrus) remains a challenge for scribal identification~\cite{d2025deep}, it forces the model to focus on diachronic morphological stability.

\section{Conclusions}
We introduce three datasets of historical Greek handwriting (Hell-Char, PaLit-Char, Med-Char), using them to examine how modern representation learning captures symbolic variation across time. Beyond establishing Hell-Char as a benchmark for low-resource, domain-shifted visual recognition, we propose two methodological innovations: a similarity-weighted supervised contrastive loss, which explicitly models structured inter-class similarity, and a lacuna-driven augmentation scheme, which simulates realistic manuscript degradations. Empirically, we demonstrate that CNN-derived embeddings yield more discriminative and structured representations than PCA or generic pre-trained features, while clustering uncovers stylistic subgroups that mirror diachronic variation and coexisting scribal conventions. Prototype visualisations per letter–century further illustrate gradual graphical change, providing interpretable bridges between computational analysis and paleographic interpretation. More broadly, our findings highlight the limits of natural-image transfer learning for historical document analysis and show that incorporating domain-informed similarity and degradation modelling improves robustness under temporal domain shift. 

\section*{Acknowledgements}
This work has been partially supported by: project MIS 5154714 of the National Recovery and Resilience Plan Greece 2.0 funded by the European Union under the NextGenerationEU Program; and the Swiss National Science Foundation, SNSF Starting Grant TMSGI1-211682 “EGRAPSA: Retracing the evolutions of handwritings in Greco-Roman Egypt thanks to digital paleography.”

\bibliography{refs}
\bibliographystyle{splncs04}

\end{document}